# Deep learning-based interactive segmentation in remote sensing

Zhe Wang[1], Shoukun Sun[2], Xiang Que[3,4 *], Xiaogang Ma[3], and Carmen Galaz García[1,5]

1. National Center for Ecological Analysis and Synthesis, University of California, Santa Barbara, Santa Barbara, 93101, USA
2. Department of Computer Science, University of Idaho, Idaho Falls, ID 83401, USA
3. Department of Computer Science, University of Idaho, Moscow, ID 83844, USA
4. College of Computer and Information Sciences, Fujian Agriculture and Forestry University, Fuzhou, 350002, China
5. Bren School of Environmental Science and Management, University of California, Santa Barbara, Santa Barbara, CA 93106, United States of America
* Corresponding Author: xiangq@uidaho.edu

Abstract:

Interactive segmentation, a computer vision technique where a user provides guidance to help an algorithm segment a feature of interest in an image, has achieved outstanding accuracy and efficient human-computer interaction. However, few studies have discussed its application to remote sensing imagery, where click-based interactive segmentation could greatly facilitate the analysis of complicated landscapes. This study aims to bridge the gap between click-based interactive segmentation and remote sensing image analysis by conducting a benchmark study on various click-based interactive segmentation models. We assessed the performance of five state-of-the-art interactive segmentation methods (Reviving Iterative Training with Mask Guidance for Interactive Segmentation (RITM), FocalClick, SimpleClick, Iterative Click Loss (ICL), and Segment Anything (SAM)) on two high-resolution aerial imagery datasets. The Cascade-Forward Refinement (CFR) approach, an innovative inference strategy for interactive segmentation, was also introduced to enhance the segmentation results without requiring manual efforts. We further integrated CFR into all models for comparison. The performance of these methods on various land cover types, different object sizes, and multiple band combinations in the datasets was evaluated. The SimpleClick-CFR model consistently outperformed the other methods in our experiments. Building upon these findings, we developed a dedicated online tool called SegMap for interactive segmentation of remote sensing data. SegMap incorporates a well-performing interactive model that is fine-tuned with remote sensing data. Unlike existing interactive segmentation tools, SegMap offers robust interactivity, modifiability, and adaptability to analyze remote sensing imagery.

## 1. Introduction

Applying deep learning techniques in remote sensing has revolutionized the analysis of complex datasets by automating feature extraction and improving object detection through powerful models like convolutional neural networks (CNNs) (Kattenborn et al., 2021). These models efficiently handle a large amount of data, achieving high accuracy in land use classification(Song et al., 2019) and change detection (Kattenborn et al., 2021; Seydi et al., 2020; Zhang et al., 2018). A critical aspect of remote sensing analysis is image segmentation, which involves distinguishing and delineating features or regions within the images. Deep learning-based image segmentation leverages neural networks to automatically identify complex patterns and features, accurately delineating objects like trees, buildings, and water bodies (Cheng et al., 2020). While fully automated deep learning-based segmentation offers significant accuracy and efficiency, integrating user input through interactive segmentation can further refine results (Hichri et al., 2012). Interactive segmentation allows users to guide and correct the algorithm's output, enhancing segmentation quality, especially in challenging scenarios where automated methods alone may struggle to achieve optimal precision (Olabarriaga and Smeulders, 2001). It allows users to guide the segmentation process by providing additional inputs such as scribbles or bounding boxes to refine and improve the segmentation results (Zhang et al., 2020). Among these interactions, click inputs can significantly reduce annotation time and costs in remote sensing by enabling precise object delineation through minimal user inputs like clicks, which is crucial for handling complex aerial imagery with small or ambiguous features (Li et al., 2024; Sun et al., 2023). However, click-based interactive segmentation in remote sensing image analysis remains underexplored.

The Segment Anything Model (SAM) has been the most discussed interactive segmentation model in the past two years. SAM is a groundbreaking foundation model for image segmentation, capable of zero-shot generalization to new tasks and image distributions (Kirillov et al., 2023). Recent research has explored its application to remote sensing imagery. Osco et al. (2023) investigated SAM's performance on aerial and orbital images, proposing a technique combining text prompts with one-shot training to improve accuracy (Osco et al., 2023). Yan et al. (2023) developed RingMo-SAM, a foundation model capable of segmenting and categorizing objects in optical and SAR remote sensing data (Yan et al., 2023). Wu & Osco (2023) introduced SAMGEO, a Python package that simplifies geospatial data segmentation using SAM, offering automatic, interactive, and text-prompt-based segmentation options (Wu and Osco, 2023). While these studies highlight the feasibility of SAM for remote sensing and emphasize the strength of its text prompts, it remains uncertain whether SAM's text prompts alone are sufficient to accurately perform complex ecological and environmental segmentation tasks, such as segmenting complex farmland and vegetation classifications or assigning specific descriptors to houses.

Click prompts can be more suitable for interactive segmentation in remote sensing than text prompts (Moskalenko et al., 2024). They allow users to specify exact locations on an image, essential for accurately delineating features such as roads, buildings, or vegetation types. Additionally, click prompts simplify the segmentation process by reducing complexity and enhancing adaptability. Given the variability in remote sensing images, ranging in resolution and content, clicks provide a straightforward way for users to focus on areas of interest without the need for precise textual descriptions. Numerous approaches have been proposed to address the challenges and limitations of click-based interactive segmentation methods. Sofiiuk et al. (2020) proposed a feature backpropagating refinement scheme (f-BRS), an interactive segmentation method that optimizes intermediate parameters to reduce computation time while improving accuracy (Sofiiuk et al., 2020). Then, his team further proposed Reviving Iterative Training with Mask Guidance for Interactive Segmentation (RITM) by replacing f-BRS's backpropagation-based refinement with RITM's iterative mask-guided training and a scaled interactive branch, enhancing both accuracy and stability (Sofiiuk et al., 2022). Chen et al. (2022) developed FocalClick, a model that improves efficiency and performance for existing masks and introduced the Vision Transformer (ViT) backbone to the architecture (Chen et al., 2022). Liu et al. (2022) proposed SimpleClick, an interactive image segmentation method that achieves good results with a simple implementation (Liu et al., 2022). Based on the SimpleClick architecture, Sun et al. (2023) introduced a Cascade-Forward Refinement (CFR) with the Iterative Click Loss (ICL) model, which consists of three components that can improve segmentation quality without increasing the number of user interactions (Sun et al., 2024).

This study aims to bridge the gap between click-based interactive segmentation and remote sensing image analysis through a benchmark of various interactive segmentation models. Our primary objectives are to:

(1) Investigate the performance of state-of-the-art click-based interactive segmentation methods on two International Society for Photogrammetry and Remote Sensing (ISPRS) datasets.

(2) Evaluate interactive segmentation methods across diverse land cover types, object sizes, and band combinations.

(3) Explore integrating the Cascade-Forward Refinement (CFR) strategy into the current click-based interactive segmentation methods, which perform well on remote sensing data.

(4) Develop a novel tool for click-based interactive segmentation tailored for remote sensing images.

## 2. Materials and methods

### 2.1 Data

In this study, we acquired the Potsdam and Vaihingen datasets from ISPRS (Figure 1). These high-resolution aerial imagery datasets are designed for semantic labeling and object detection tasks ("2D Semantic Labeling - Potsdam," n.d.; "Vaihingen 2D Semantic Labeling - ISPRS," n.d.). The Potsdam datasets contain 38 images, each with spectral bands red (R), green (G), blue (B), and near-infrared (IR). We used 32 for training for each dataset and reserved 6 for testing. Each image has a resolution of 6000×6000 pixels and a spatial resolution of 5 cm. The Vaihingen datasets contain 33 images, each with R, G, and IR. We used 27 for training and reserved 6 for testing. Each Vaihingen image has a resolution of approximately 2000×2500 pixels with a spatial resolution of 9 cm.

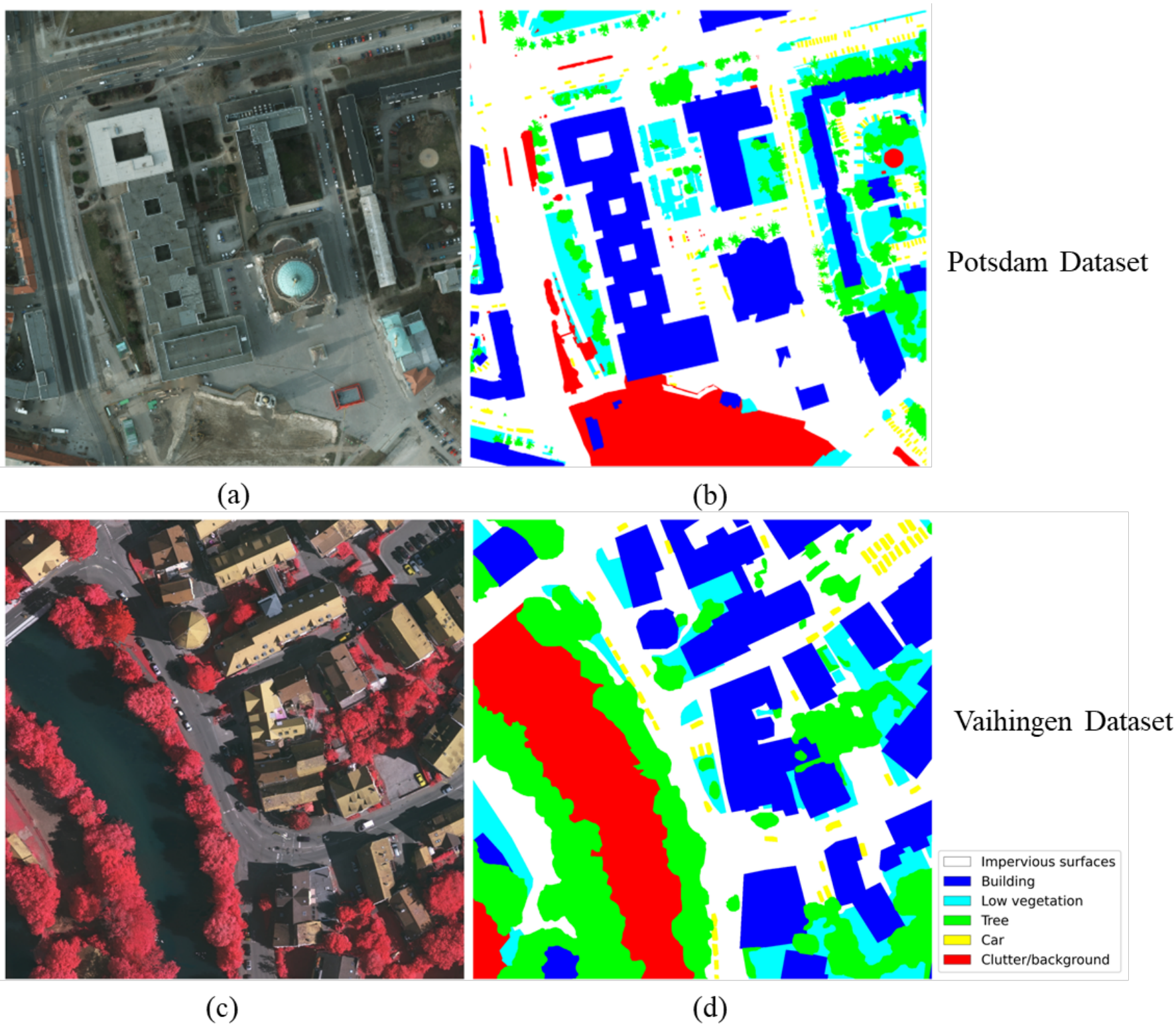

Figure 1. Example patches of Potsdam Dataset (RGB image (a) and corresponding labels (b)) and Vaihingen Dataset (RGIR image (c) and corresponding labels (d))

Figure 2 illustrates example patches featuring RGB imaging, RGIR imaging, and individual imaging across four bands (R, G, B, IR) within the PRGBIR dataset. An RGIR composite image in remote sensing can help make the vegetation appear more obvious and more differentiated from other features. This is because it utilizes the near-infrared band, a spectrum to which vegetation strongly responds (Uchegbulam and Ayolabi, 2013). Hence, vegetation typically appears bright red when viewed in an RGIR composite image.

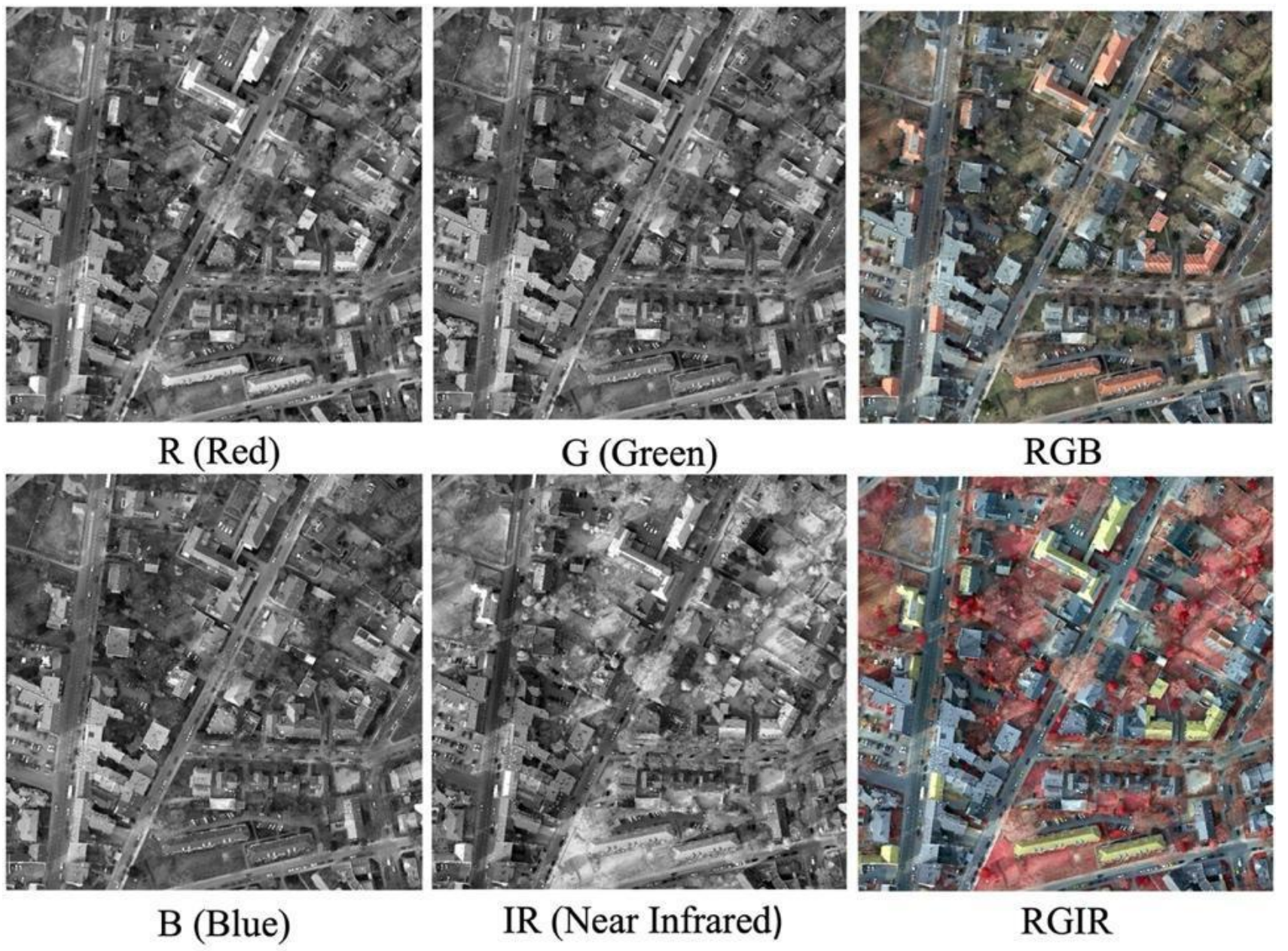

Figure 2. Comparative visualization of RGB, RGIR, R, G, B, and IR imaging

### 2.2 Data preprocessing

Based on the two datasets, we remade them into four datasets with different band combinations, PRGB (Potsdam RGB), PRGIR (Potsdam RGIR), PRGBIR (Potsdam RGBIR), and VRGIR (Vaihingen RGIR). (where infrared data is often coded in red), making it easier to identify and examine.

We further refined the four datasets into instance segmentation datasets. This was achieved by dividing each image into patches and converting every continuous label of a specific class within a patch into an object. The Potsdam dataset was split into 600x600 pixel patches, resulting in 3200 patches and 47152 objects for training, and 600 patches and 10365 objects for testing. The Vaihingen dataset was processed similarly, resulting in 1770 patches and 20102 objects for training and 385 patches and 4651 objects for testing.

### 2.3 Representable models

Interactive segmentation methods often require interaction, like clicks, bounding boxes, freehand, and scribbles (Pavoni et al., 2022). Click-based interaction has several practical advantages in interactive segmentation, offering precision and efficiency. The specific localization of points allows users to pinpoint areas of interest accurately, resulting in precise segmentation. It's efficient due to minimal interaction cost, requiring only a few clicks or keypresses. This makes it ideal for users to review and refine segmentation results in a feedback loop, allowing prompt corrections or improvements.

#### 2.3.1 Reviving Iterative Training with Mask Guidance for Interactive Segmentation (RITM) (Sofiiuk et al., 2022)

RITM uses a feedforward model for click-based segmentation, enabling interactive segmentation without additional optimization schemes. It can segment new objects and refine existing masks iteratively. The model leverages segmentation masks from previous steps to enhance accuracy and stability, achieving state-of-the-art results without the need for backward passes during inference. The training dataset choice significantly impacts segmentation quality, with a combination of the COCO and LVIS datasets (Gupta et al., 2019; Lin et al., 2014) showing superior performance. This method emphasizes simplicity and effectiveness in interactive segmentation tasks.

#### 2.3.2 FocalClick (Chen et al., 2022)

FocalClick addresses the challenges of interactive segmentation by introducing an efficient pipeline that decomposes the heavy inference task into two lighter predictions on smaller patches. Unlike some earlier models, which might rely strictly on traditional Convolutional Neural Networks (CNNs) or don't utilize the previous masks for iterative refinement, FocalClick incorporates both CNN and ViT architectures. This dual-architecture approach allows FocalClick to leverage the spatial hierarchies captured by CNNs and the global context understanding facilitated by ViTs, providing more nuanced and accurate segmentation outputs. FocalClick also distinguishes itself by utilizing previous masks as part of its input for refinement, suggesting an iterative, feedback-driven approach to improve segmentation accuracy. This feature allows the system to progressively refine the segmentation output based on user inputs, making the tool more interactive and user-friendly, especially in complex segmentation tasks where precision is crucial.

#### 2.3.3 SimpleClick (Liu et al., 2022)

SimpleClick introduces a novel approach to interactive segmentation by leveraging only a plain ViT backbone. The model utilizes a masked autoencoder (MAE) pretraining strategy to capture meaningful representations efficiently. By adapting the ViT backbone for interactive segmentation with minimal modifications, SimpleClick achieves state-of-the-art performance in segmentation tasks. The deep neural network architecture divides input images into fixed-size patches, which are linearly projected into vectors and processed through a series of Transformer blocks for self-attention.

#### 2.3.4 ICL-CFR (Sun et al., 2023)

Based on SimpleClick, the CFR-based inference strategy is central to the proposed interactive image segmentation framework, enabling iterative refinement of segmentation masks during inference.

The CFR strategy is defined by:

$$Y_0^t = f(X, P^t, Y_n^{t-1}) \tag{1}$$

$$Y_i^t = f\left(X, P^t, Y_{i-1}^t\right), i \in \{1,2,3, \ldots, n\} \tag{2}$$

where $Y_0^t$ represents the output of the $t$-$th$ coarse segmentation step (outer loop). It's the initial segmentation result at the current stage. $f$ is the function or model being used to generate the segmentation. It processes the inputs to produce segmentation masks. $X$ is the raw image input that is being segmented. $P^t$ is the set of user clicks at step $t$ . These are user interactions collected at the $t$-th step to guide the segmentation. $Y_{i-1}^t$ is the refined mask from the previous step $(t-1)$. It provides the model with context from the last refined result.

The CFR process with two loops, the outer loop and the inner loop. It begins with a coarse segmentation step, where an initial mask $Y_0^t$ is generated using the function $f$ based on the raw image $X$, user-generated click maps $P^t$, and the refined mask from the previous step $Y_n^{t-1}$. This forms the outer loop. Then, in the inner loop, the method refines this segmentation iteratively through successive steps using Equation (2), where each stage $Y_i^t$ for $i \in \{1,2,3, \dots, n\}$, builds upon the mask obtained from the prior iteration, $Y_{i-1}^t$. The number of refinement steps $n$ can be predetermined or adaptively set, enabling continuous enhancement of segmentation accuracy.

The outer loop handles user interactions by updating the segmentation mask with each new click, while the inner loop refines the mask automatically through multiple model passes without additional user input. This iterative approach allows the segmentation mask to be progressively refined without requiring additional user interactions, leveraging both the initial input and subsequent refinements for high-quality outputs.

**2.3.5 SAM (Kirillov et al., 2023)**

The SAM developed by Meta is an advanced segmentation model capable of accurately segmenting any object within an image using just a single click. SAM is a promotable segmentation system that can generalize its segmentation capabilities to unfamiliar objects and images without additional training. SAM's exceptional performance is attributed to its extensive training on a large dataset of 11 million images and 1.1 billion masks. With this comprehensive training, SAM can effectively generate masks for all objects in an image. Alternatively, users can provide specific points or guidance to SAM, enabling the model to create precise masks tailored to a particular object of interest. Table 1 outlines the evolution and capabilities of the five interactive segmentation models evaluated in this study.

Table 1. Comparison between different models

| **Models** | **Input** | | | **Refinement** | **Backbone** | | **Year** |
|---|---|---|---|---|---|---|---|
| | Image | Clicks | Previous Mask | | CNN | ViT | |
| **RITM** | ✓ | ✓ | × | × | ✓ | × | 2021 |
| **FocalClick** | ✓ | ✓ | ✓ | ✓ | ✓ | ✓ | 2022 |
| **SimpleClick** | ✓ | ✓ | ✓ | × | × | ✓ | 2022 |
| **ICL-CFR** | ✓ | ✓ | ✓ | ✓ | × | ✓ | 2023 |
| **SAM** | ✓ | ✓ | ✓ | × | × | ✓ | 2023 |

## 3. Experiments and results

Our experiments evaluated five interactive segmentation models (listed in section 2.3) across three steps. First, we tested eight variants: pre-trained versions of SimpleClick, ICL, RITM, FocalClick, and SAM, along with CFR-enhanced iterations of SimpleClick, ICL, and RITM (FocalClick excluded due to its incompatible local refinement module). Initial evaluations revealed that SAM and FocalClick underperformed, prompting a second-step comparison of ICL, RITM, SimpleClick, and their CFR versions on the PRGB, PRGIR, and VRGIR datasets. In the third step, we addressed the mismatch between the three-channel model inputs and the four-band PRGBIR dataset by testing band combinations and land cover scenarios. Ultimately, SimpleClick emerged as the top performer, leading us to adapt its architecture to process all four bands from the PRGBIR dataset. Table 2 summarizes the experimental setup and results for each model.

Table 2. Experiments of each model on different datasets

| Model Name | Pretrained | PRGB | PRGIR | VRGIR | PRGBIR |
|---|---|---|---|---|---|
| SimpleClick | ✓ | ✓ | ✓ | ✓ | ✓ |
| SimpleClick-CFR | ✓ | ✓ | ✓ | ✓ | ✓ |
| ICL | ✓ | ✓ | ✓ | ✓ | × |
| ICL-CFR | ✓ | ✓ | ✓ | ✓ | × |
| RITM | ✓ | ✓ | ✓ | ✓ | × |
| RITM-CFR | ✓ | ✓ | ✓ | ✓ | × |
| FocalClick | ✓ | × | × | × | × |
| SAM | ✓ | × | × | × | × |

### 3.1 Evaluation of models with initial weights

We evaluated model performance using the number of clicks (NoC) (Sofiiuk et al., 2020), which is the average click number required to reach the target intersection over the union (IoU) (Nowozin, 2014). These metrics quantify clicks or user interactions needed to achieve a desired result. Lower NoC suggests a more efficient model as it requires fewer user interactions for accurate segmentation. The average number of clicks needed to reach the target IoU can be found using iterative interaction and segmentation mechanisms. Initially, the model might generate an initial segmentation with low IoU. Then, in the subsequent iterations, additional user clicks can be introduced to improve the segmentation until the target IoU is reached. We report NoC values at four IoU thresholds, 80%, 85%, 90%, and 95%, denoted as NoC80, NoC85, NoC90, and NoC95, respectively.

Table 3 shows the performance of different interactive segmentation models on PRGB and PRGIR datasets without fine-tuning at varying IoU levels. On the PRGB dataset, RITM demonstrates the best performance at the 80% IoU level. The SimpleClick-CFR model performs best at the higher IoU levels (85%, 90%, and 95%). Except for SAM and Focal Click, the remaining models exhibit similar performance at the higher IoU levels (85%, 90%, and 95%). However, as the desired IoU level increases, the efficiency of all models declines significantly. While the least-performing model at the 80% IoU level requires fewer than seven clicks, at the 95% IoU level, even the best-performing model, SimpleClick-CFR, requires approximately fourteen clicks. This trade-off between precision and efficiency becomes evident at higher IoU thresholds. On the PRGIR dataset, SimpleClick and SimpleClick-CFR consistently demonstrate strong performance across multiple IoU thresholds. SimpleClick requires the fewest clicks at 80% IoU and maintains its superiority at 85%, 90%, and 95% IoU levels. RITM and RITM-CFR models also exhibit competitive performance across different IoU thresholds. However, SAM consistently requires the highest number of clicks.

Table 3. Evaluation results on PRGB and PRGIR datasets without fine-tuning

| **Model Name** | **PRGB** | | | | **PRGIR** | | | |
|---|---|---|---|---|---|---|---|---|
| | **NoC80** | **NoC85** | **NoC90** | **NoC95** | **NoC80** | **NoC85** | **NoC90** | **NoC95** |
| SimpleClick | 4.90 | 6.27 | 8.84 | 14.31 | **4.83** | 6.45 | **9.24** | 14.95 |
| SimpleClick-CFR | 4.94 | **6.26** | **8.80** | **14.28** | 4.90 | **6.44** | 9.29 | **14.83** |
| ICL | 5.04 | 6.47 | 9.10 | 14.61 | 5.10 | 6.65 | 9.64 | 15.24 |
| ICL-CFR | 5.14 | 6.54 | 9.15 | 14.62 | 5.11 | 6.73 | 9.63 | 15.20 |
| RITM | **4.78** | 6.34 | 9.22 | 14.89 | 5.22 | 6.87 | 10.00 | 15.52 |
| RITM-CFR | 4.81 | 6.34 | 9.23 | 14.90 | 5.36 | 7.08 | 10.14 | 15.53 |
| FocalClick | 5.56 | 7.05 | 9.70 | 15.01 | 5.26 | 6.96 | 10.04 | 15.28 |
| SAM | 6.58 | 8.77 | 12.34 | 17.54 | 6.78 | 9.34 | 13.36 | 17.78 |

Figure 3 reveals the varying trends of mean intersection over the union (mIoU) for eight distinct models as the number of clicks increases from 1 to 10. The RITM-CFR model is the most effective at the outset with a single click. As the click count rises from 2 to 4, the performance of the SAM model remains consistent and comparable to both RITM and RITM-CFR. Yet, once the click count surpasses 5, SAM persistently registers a slight underperformance relative to the other models. Upon reaching 10 clicks, all models display a mIoU near 0.9, except SAM, which falls below 0.85.

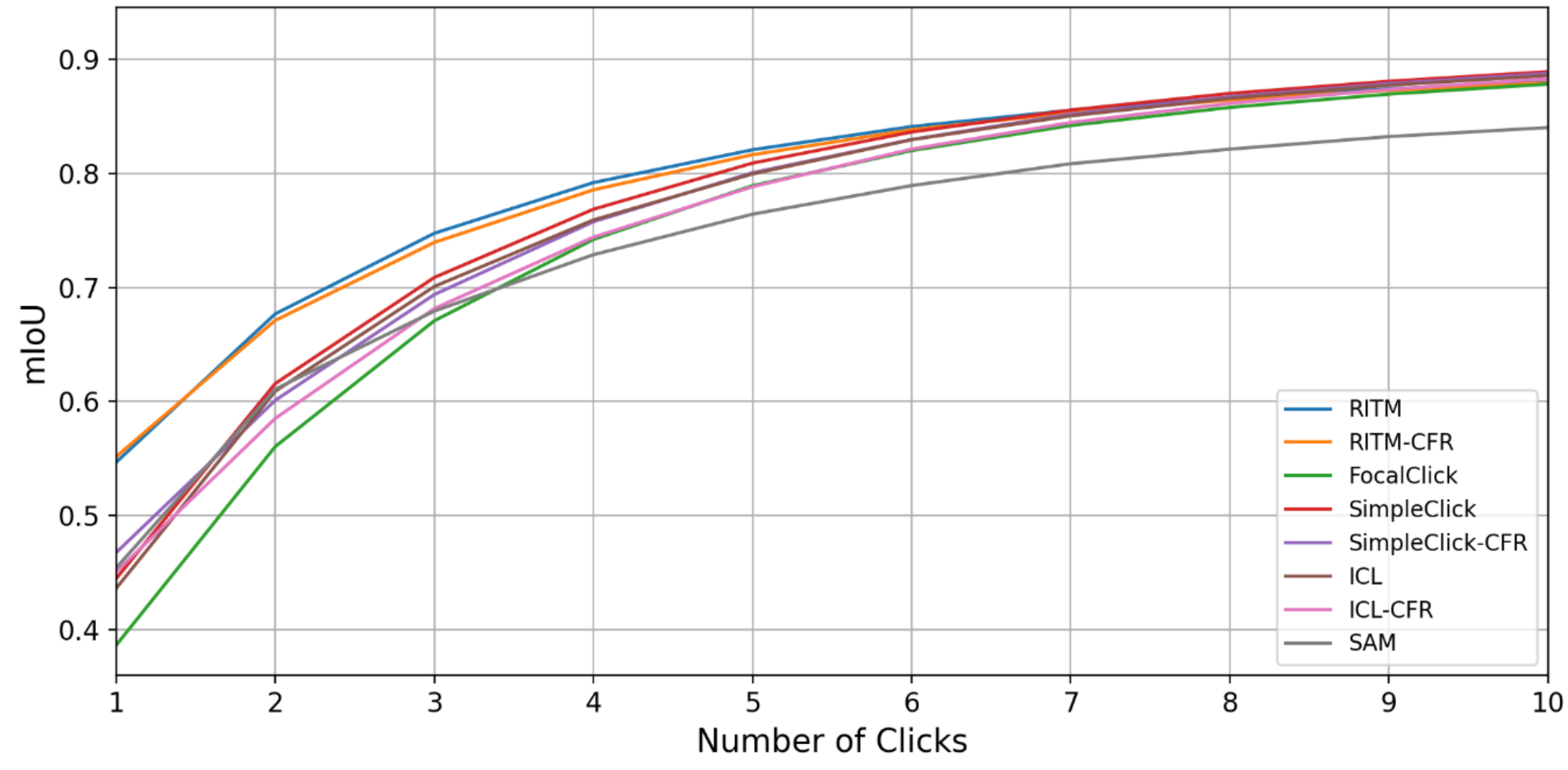

Figure 3. Mean IoU (mIoU) given *n* clicks (*n* ranges from 1 to 10) for the RITM, RITM, … models.

### 3.2 Model evaluations after fine-tuning

Table 4 shows the results of interactive segmentation models after this fine-tuning is applied to segmenting the PRGB and PRGIR datasets. SimpleClick and SimpleClick-CFR consistently exhibit superior performance with lower click counts across all IoU thresholds, indicating their remarkable efficiency in user interaction for accurate segmentation. The ICL and ICL-CFR models exhibit slightly higher click counts, indicating increased user interaction. However, they still deliver reasonable performance with a moderate number of clicks to achieve accurate segmentation results. Similarly, the RITM and RITM-CFR models present slightly higher click counts but maintain efficient user interaction to achieve precise segmentation.

Table 4. Evaluation results on PRGB and PRGIR datasets after finetuning

| Model Name (Finetuned) | PRGB | | | | PRGIR | | | |
|---|---|---|---|---|---|---|---|---|
| | NoC80 | NoC85 | NoC90 | NoC95 | NoC80 | NoC85 | NoC90 | NoC95 |
| SimpleClick | 3.26 | 4.39 | 6.73 | 12.47 | 3.30 | 4.42 | 6.77 | 12.45 |
| SimpleClick-CFR | 3.28 | 4.39 | 6.74 | 12.45 | 3.27 | 4.39 | 6.77 | 12.43 |
| ICL | 3.70 | 5.04 | 7.69 | 13.61 | 4.32 | 5.87 | 8.90 | 15.01 |
| ICL-CFR | 3.64 | 4.95 | 7.59 | 13.55 | 4.16 | 5.73 | 8.74 | 14.93 |
| RITM | 3.92 | 5.36 | 8.18 | 14.10 | 4.18 | 5.67 | 8.60 | 14.45 |
| RITM-CFR | 3.86 | 5.30 | 8.13 | 14.03 | 4.16 | 5.62 | 8.54 | 14.36 |

Figure 4 provides a comparative analysis of the mIoU for models before and after fine-tuning, as the number of clicks varies from 1 to 10. With only one click, fine-tuned models already achieve a mIoU exceeding 0.6. As the click count progressively increases, the fine-tuned models consistently outperform their non-fine-tuned counterparts. However, the performance gap, as quantified by the mIoU, between the two types of models steadily narrows. By the point where the number of click counts hits 10, the superior performance of the finetuned models, though still discernible, approximates that of the non-fine-tuned models, with scores hovering around 0.9.

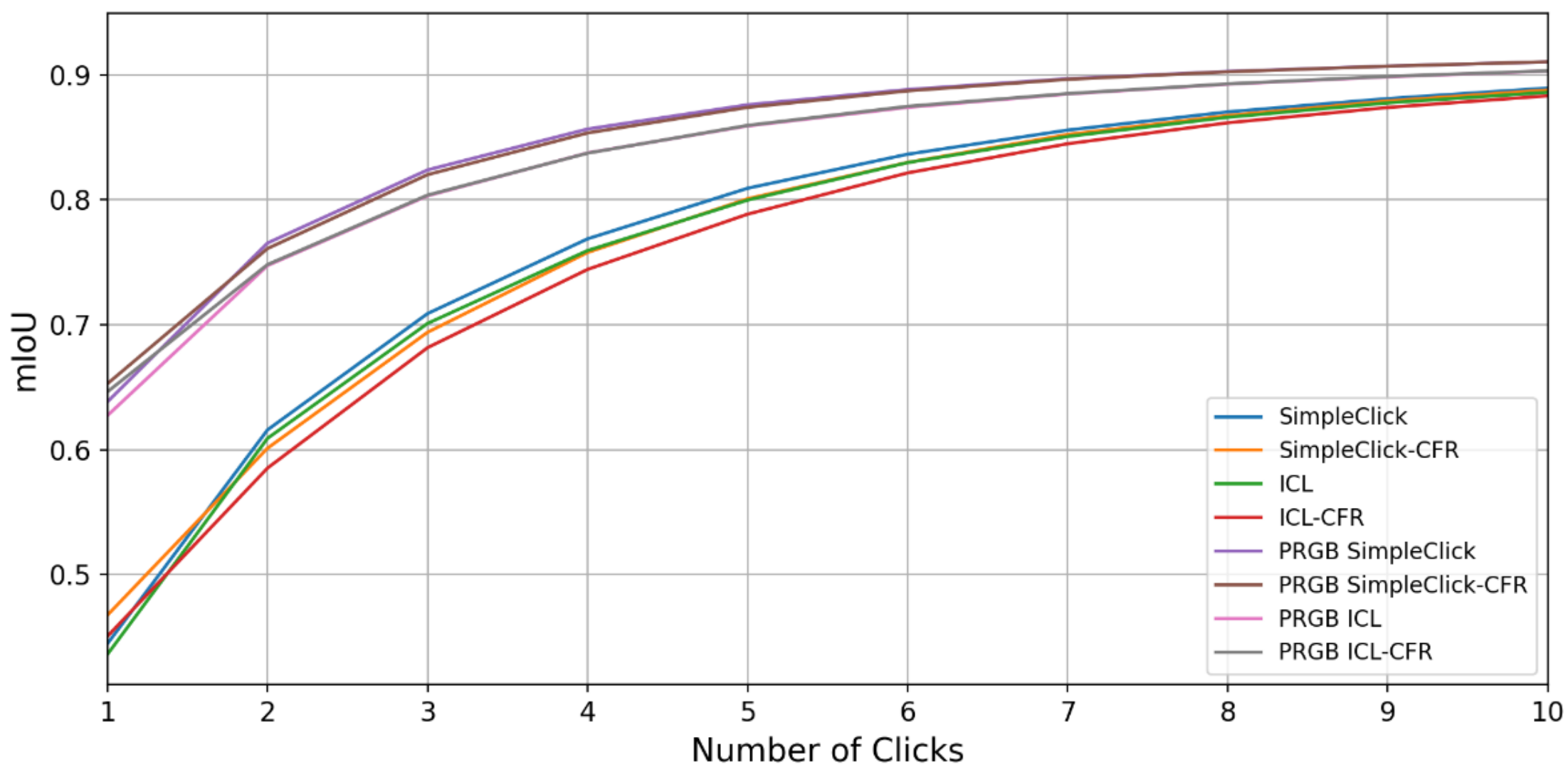

Figure 4. Comparison of mean IoU (mIoU) given *n* clicks (*n* ranges from 1 to 10) between models before and after fine-tuning.

Table 5 demonstrates the evaluation results on the VRGIR dataset before and after fine-tuning with the PRGIR dataset. Among the various models we evaluated, the SimpleClick models consistently outperformed the others, achieving superior results across all IoU levels while requiring fewer user clicks for interactive segmentation, except the ICL and ICL-CFR models. Notably, the SimpleClick-CFR model demonstrates the highest level of improvement, exhibiting a remarkable increase of 17.79% at the 80% IoU level. These findings highlight the robustness and efficiency of the SimpleClick model in the context of multispectral data. Moreover, comparing all the models to the baseline model that was not fine-tuned and directly evaluated on the VRGIR dataset, we observed significant improvements in performance across the board.

Table 5. Evaluation results on the VRGIR dataset before and after fine-tuning

| **Model Name (Finetuned)** | **Before fine-tuning** | | | | **After fine-tuning** | | | |
|---|---|---|---|---|---|---|---|---|
| | **NoC80** | **NoC85** | **NoC90** | **NoC95** | **NoC80** | **NoC85** | **NoC90** | **NoC95** |
| SimpleClick | 4.83 | 6.45 | 9.24 | 14.95 | **4.16** | 5.7 | 8.43 | 14.19 |
| SimpleClick-CFR | 4.9 | 6.44 | 9.29 | 14.83 | 4.16 | 5.67 | 8.34 | 14.22 |
| ICL | 5.1 | 6.65 | 9.64 | 15.24 | 5.24 | 6.95 | 10.16 | 15.72 |
| ICL-CFR | 5.11 | 6.73 | 9.63 | 15.2 | 5.17 | 6.92 | 10.21 | 15.73 |
| RITM | 5.22 | 6.87 | 10 | 15.52 | 4.81 | 6.6 | 9.65 | 15.19 |
| RITM-CFR | 5.36 | 7.08 | 10.14 | 15.53 | 4.84 | 6.47 | 9.62 | 15.15 |

### 3.3 Model performance by different classes

Tables 6 and 7 present the results of the interactive segmentation models segmenting the classes ‘Impervious Surface’, ‘Building’, ‘Low Vegetation’, ‘Tree’, and ‘Car’. Given that all models exhibit a similar trend in their results, our analysis focuses on evaluating their performance after fine-tuning them with the PRGB dataset. In the SimpleClick model, we observed differing levels of performance across classes.

In Table 6, the 'Building' class achieved the lowest number of clicks required for accurate segmentation. Conversely, the 'Low Vegetation' class presented a more significant challenge. The 'Tree' class required an intermediate number of clicks, while the 'Car' class required more clicks for successful segmentation. Similar trends were also observed in the SimpleClick-CFR model, with the 'Building' class demonstrating the lowest number of clicks (1.83) required for an 80% IoU. For the 'Low Vegetation' class, 4.17 clicks were necessary, while the 'Tree' class needed 3.27 clicks, and the 'Car' class required 1.79 clicks for similar IoU levels.

Table 6. Evaluation results of different land cover types using the PRGB dataset for testing.

| **Class** / **Model Name (Finetuned)** | **Impervious surface** | **Building** | **Low vegetation** | **Tree** | **Car** |
|---|---|---|---|---|---|
| | **Noc80/95** | **Noc80/95** | **Noc80/95** | **Noc80/95** | **Noc80/95** |
| SimpleClick | 3.53/12.33 | **1.85/4.63** | 4.11/13.49 | 3.22/12.45 | 1.76/10.45 |
| SimpleClick-CFR | 3.55/12.28 | **1.83/4.69** | 4.17/13.47 | 3.27/12.36 | 1.79/10.46 |
| ICL | 4.07/13.67 | 1.96/5.10 | 4.77/14.70 | 3.62/14.16 | 1.99/11.67 |
| ICL-CFR | 3.97/13.62 | 1.99/5.04 | 4.70/14.64 | 3.57/14.07 | 1.98/11.58 |
| RITM | 4.46/14.62 | 2.30/6.14 | 4.89/15.22 | 4.11/14.55 | 2.09/11.95 |
| RITM-CFR | 4.42/14.48 | 2.31/6.20 | 4.82/15.07 | 4.01/14.58 | 2.01/11.86 |

Table 7 shows the performance of different interactive segmentation models after finetuning the PRGB dataset. The observed trend in this table aligns with the patterns observed in Table 4, indicating consistency in the model performances across the evaluated datasets. One intriguing observation is that despite the expected benefits of utilizing the IR band for vegetation segmentation (e.g. trees and grass), the results in Table 7 indicate lower efficiency when using the RGIR bands for most classes compared ton using RGB bands (Table 6). This suggests that including the IR does not always enhance segmentation performance.

Consistent with the observations in Table 1, achieving higher IoU levels, particularly 95%, requires significantly more clicks. The findings show that most classes require approximately three times more clicks to reach the desired IoU level from 80% to 95%. Notably, the 'Car' class stands out, necessitating around nine times more clicks. For example, in the SimpleClick model, the click counts increase from 1.85 to 10.40.

Table 7. Evaluation results of different land cover types using the PRGIR dataset for testing.

| **Class** / **Model Name** | Impervious surface | Building | Low vegetation | Tree | Car |
|---|---|---|---|---|---|
| | Noc80/95 | Noc80/95 | Noc80/95 | Noc80/95 | Noc80/95 |
| SimpleClick | **3.55/12.25** | 1.79/4.68 | **4.17/13.51** | 3.14/12.30 | 1.85/10.40 |
| SimpleClick-CFR | 3.55/12.24 | **1.78/4.67** | 4.18/13.42 | **3.12/12.27** | **1.83/10.48** |
| ICL | 4.72/15.10 | 2.29/6.09 | 5.56/15.78 | 4.01/15.56 | 2.36/14.54 |
| ICL-CFR | 4.56/15.11 | 2.23/5.89 | 5.40/15.63 | 3.86/15.55 | 2.29/14.41 |
| RITM | 4.62/15.00 | 2.51/6.58 | 5.33/15.47 | 4.47/15.13 | 2.19/12.38 |
| RITM-CFR | 4.68/14.92 | 2.52/6.63 | 5.30/15.38 | 4.49/15.09 | 2.16/12.20 |

### 3.4 Model performance on different segment sizes

The size of surface objects is crucial to land cover feature extraction (Weng, 2012). We further investigated the association between the size of surface objects and the number of clicks required to achieve various IoU thresholds in interactive segmentation. Experiments on interactive segmentation considered five land cover types, among which the shapes of buildings and trees exhibit relative regularity and variations compared to the other classes. Therefore, we analyze the performance of different interactive segmentation models specifically for buildings and trees of varying sizes.

To explore the performance of the models relative to building size, the buildings were grouped into five categories ranging from 0.075 $m^2$ to 881.90 $m^2$ using the quantile method. Table 8 shows the results of the classification. Larger building segments require fewer clicks to achieve satisfactory IoU levels across the interactive segmentation models. Interestingly, when comparing the performance of building with other land cover classes, we observed that buildings require fewer clicks for satisfactory segmentation. This is likely due to their regular and well-defined shape characteristics, which facilitate the interactive segmentation process, reducing click requirements for different segment sizes.

Table 8. Evaluation results of buildings with different segment sizes on the PRGB dataset

| **Size** / **Model Name (Finetuned)** | (0.075 $m^2$, 9.03 $m^2$][a] | (9.03 $m^2$, 36.36 $m^2$] | (36.36 $m^2$, 100.08 $m^2$] | (100.08 $m^2$, 256.83 $m^2$] | (256.83 $m^2$, 881.90 $m^2$] |
|---|---|---|---|---|---|
| | Noc80/95 | Noc80/95 | Noc80/95 | Noc80/95 | Noc80/95 |
| SimpleClick | **3.76/12.11** | 1.483.78 | **1.36/2.92** | 1.36/2.25 | **1.28/2.06** |
| SimpleClick-CFR | 3.62/12.36 | **1.47/3.87** | 1.37/2.88 | **1.38/2.28** | 1.30/2.03 |
| ICL | 3.91/12.48 | 1.63/4.53 | 1.41/3.38 | 1.38/2.65 | 1.44/2.42 |
| ICL-CFR | 3.99/12.37 | 1.63/4.54 | 1.41/3.27 | 1.43/2.56 | 1.48/2.44 |
| RITM | 4.09/13.99 | 1.60/4.76 | 1.50/3.93 | 1.68/3.29 | 2.64/4.72 |
| RITM-CFR | 3.83/13.78 | 1.62/4.76 | 1.53/4.20 | 1.80/3.41 | 2.76/4.81 |

a. (0.075 $m^2$, 9.03 $m^2$] indicates an interval that includes all segment sizes greater than 0.075 $m^2$ up to and including 9.03 $m^2$. The parenthesis '()' signifies that the number immediately following it is not included in the interval, while the brackets '[]' signify that the number immediately before it is included.

Table 9 outlines the results for different segment sizes of the 'Tree' class, grouping trees into five categories using the quantile method, ranging from 0.075 $m^2$ to 847.53 $m^2$. Smaller land objects require more clicks to achieve accurate segmentation, a pattern consistent across all evaluated models. Remarkably, the optimal performance in terms of click efficiency is observed within a specific range of sizes, with the best results obtained for land objects ranging from 155.93 $m^2$ to 847.53 $m^2$. This suggests that interactive segmentation models achieve a favorable balance between accuracy and the number of required clicks within this size range, resulting in more efficient and reliable segmentation outcomes.

Table 9. Evaluation results of the Tree class with different segment sizes on the PRGB Dataset

| Size / **Model Name** | (0.075 $m^2$, 1.96 $m^2$][a] | (1.96 $m^2$, 9.52 $m^2$] | (9.52 $m^2$, 37.12 $m^2$] | (37.12 $m^2$, 155.93 $m^2$] | (155.93 $m^2$, 847.53 $m^2$] |
|---|---|---|---|---|---|
| | Noc80/95 | Noc80/95 | Noc80/95 | Noc80/95 | Noc80/95 |
| SimpleClick | **3.55/12.25** | 1.79/4.68 | **4.17/13.51** | 3.14/12.30 | 1.85/10.40 |
| SimpleClick-CFR | 3.55/12.24 | **1.78/4.67** | 4.18/13.42 | **3.12/12.27** | **1.83/10.48** |
| ICL | 4.72/15.10 | 2.29/6.09 | 5.56/15.78 | 4.01/15.56 | 2.36/14.54 |
| ICL-CFR | 4.56/15.11 | 2.23/5.89 | 5.40/15.63 | 3.86/15.55 | 2.29/14.41 |
| RITM | 4.62/15.00 | 2.51/6.58 | 5.33/15.47 | 4.47/15.13 | 2.19/12.38 |
| RITM-CFR | 4.68/14.92 | 2.52/6.63 | 5.30/15.38 | 4.49/15.09 | 2.16/12.20 |

a. (0.075 $m^2$, 1.96 $m^2$] indicates an interval that includes all segment sizes greater than 0.075 $m^2$ up to and including 1.96 $m^2$. The parenthesis '()' signifies that the number immediately following it is not included in the interval, while the brackets '[]' signify that the number immediately before it is included.

### 3.5 Performance evaluation with the four-band dataset (PRGBIR)

Table 10 shows the models' performance on the PRGBIR dataset, which consists of four bands and is expected to provide more spectral information than the three-band input. However, the results indicate that the models trained on the PRGBIR dataset perform worse than those trained on three-band inputs (Tables 6 and 7). This suggests that incorporating an additional band does not necessarily enhance segmentation performance.

Table 10. Evaluation results on PRGBIR dataset

| **Model** / **Class** | **SimpleClick** | **SimpleClick-CFR** |
|---|---|---|
| | **Noc80/85/90/95** | **Noc80/85/90/95** |
| Impervious Surface | 3.96/5.06/7.50/13.40 | 3.88/4.98/7.43/13.37 |
| Building | 2.00/2.37/3.06/5.02 | 1.93/2.27/2.93/4.91 |
| Low Vegetation | 4.65/6.11/8.80/14.13 | 4.60/5.95/8.65/14.02 |
| Tree | 3.75/4.83/7.09/13.06 | 3.76/4.82/6.98/13.05 |
| Car | 2.03/2.64/4.15/11.18 | 1.95/2.50/3.83/10.88 |
| Overall | 3.72/4.97/7.46/13.11 | 3.65/4.83/7.30/13.02 |

## 4. SegMap: Deep learning-based Interactive segmentation plug-in for remote sensing data

Drawing on the findings from the above experiments, we developed SegMap, a QGIS plug-in tool for interactive segmentation tailored to remote sensing imagery. This tool integrates models fine-tuned with remote sensing data from our earlier experiments. Figure 5 illustrates the interactive segmentation process of delineating land cover objects in SegMap. A SegMap user would utilize two types of clicks for segmentation: positive and negative. Positive clicks are placed directly on the target land cover segment to indicate the desired area for inclusion. Conversely, negative clicks are positioned on regions considered irrelevant or not belonging to the intended segment, serving as markers for exclusion (Figure 5). To maintain the balance between efficiency and accuracy in line with our experimental findings, it is advisable to exercise caution and stay within 20 clicks for a single land cover object. Exceeding this threshold could decrease the quality and efficiency of segmentation. By following this guideline, users can optimize both accuracy and processing time. It is now fully open-source and has a more detailed introduction and installation process available at 'https://github.com/TitorX/SegMap-QGIS'.

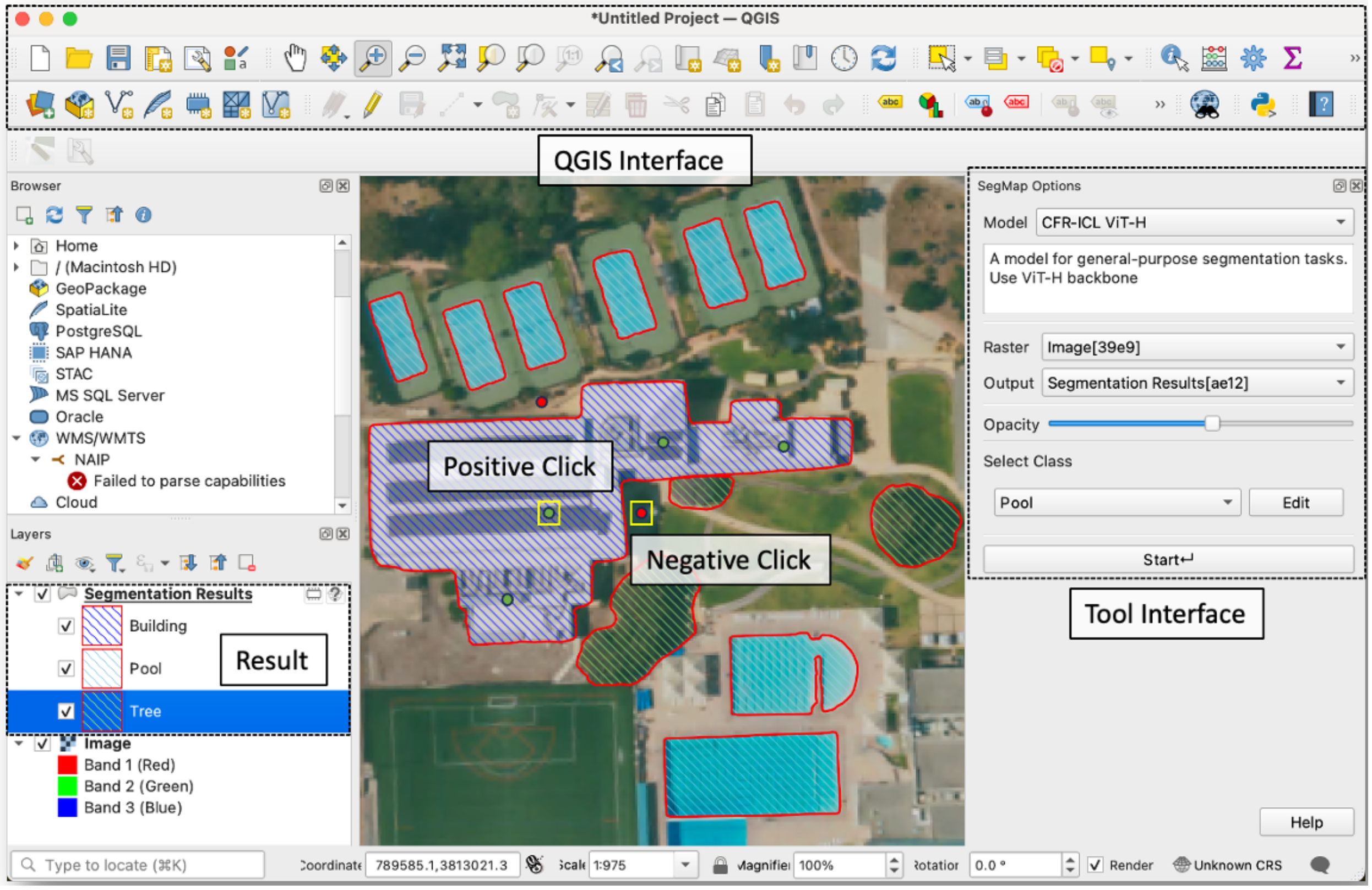

Figure 5. The SegMap interface under QGIS

Compared to other tools, such as SAM's plug-in, our tool is better suited for complex segmentation environments. Its advanced algorithms and user-friendly interface make it an ideal choice for tackling intricate segmentation tasks efficiently. Figure 6 presents an example result achieved using the click-based tool, SegMap. We selected a region within the Jack and Laura Dangermond Preserve in California, characterized by its complex and diverse vegetation. The data is from the 2020 National Agriculture Imagery Program (NAIP) imagery, with a spatial resolution of 60 cm. Leveraging existing expert knowledge, we employed the developed SegMap tool with a few points input to segment and classify the patch into seven classes: Iceplant, mixed Iceplant, Possible Iceplant, brownheaded Rush Seeps, Menzie's Golden Bush Scrub, Salt Grass Flats, and other vegetation.

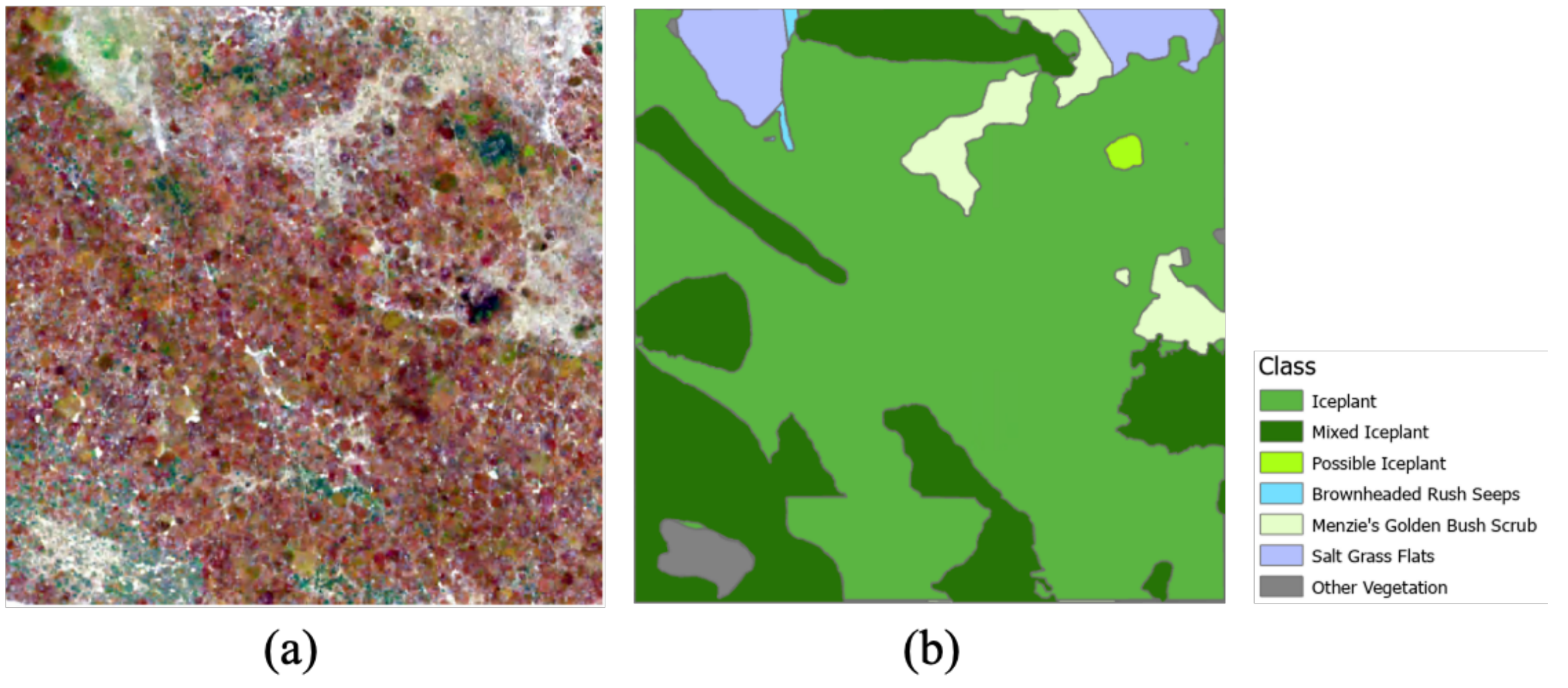

Figure 6. Example segmentation using the NAIP data in 2020 at Jack and Laura Dangermond Preserve in California: (a) NAIP RGB image and (b) corresponding segmentation result.

## 5. Discussion

### 5.1 Click-based Interactive Segmentation

Deep learning techniques have significantly improved remote sensing data analysis in the era of rapid technological evolution. Yet, the necessity of achieving high precision in image classification has brought about challenges. Despite its strength in handling massive datasets and automating the classification process, deep learning often requires labor-intensive validation or fieldwork, which questions its efficiency and accuracy. In response to these challenges, interactive segmentation emerges as a solution that synergizes expert knowledge with machine efficiency.

Interactive segmentation requires minimal human input, such as drawing boundaries or marking areas of interest in an image, which an algorithm then uses to improve segmentation results. Among the different inputs, the click-based interactive segmentation is particularly well-suited for remote sensing imagery due to its ability to incorporate user input for precise object delineation in complex environments. Remote sensing images often contain heterogeneous and intricately detailed scenes, such as various land cover types and vegetation structures, where automated segmentation might struggle to achieve high accuracy. The interactive aspect allows users to provide crucial contextual information and guidance, refining segmentation results and enhancing classification accuracy. Additionally, the click-based approach is intuitive and requires minimal training, making it accessible for experts to segment and analyze remote sensing data efficiently. This adaptability and precision make click-based interactive segmentation an effective tool for handling the unique challenges of remote sensing imagery.

Although text input or zero-shot models, such as SAM, are currently compelling, their direct application for image segmentation in remote sensing contexts still presents challenges and limitations. This is mainly due to remote sensing data's inherent complexity and diversity, which often includes various resolutions, multiple spectral bands, and feature types that these models may not be initially tailored to handle.

Additionally, the unique semantic and contextual information crucial for interpreting remote sensing imagery is often complex for these models to infer without specific domain adaptation or training.

In Figure 6, click-based interactive segmentation is well-suited for segmenting complex landscapes in remote sensing images such as those in natural preserves. This area, rich with diverse vegetation types, includes species like iceplant, brownheaded Rush Seeps, and Menzie's Golden Bush Scrub, which require specific background knowledge for accurate classification. The adaptability of click-based segmentation allows experts to interactively provide input, leveraging their domain expertise to achieve improved outcomes. This method reduces the likelihood of misclassification, particularly for rare vegetation, thus enhancing the reliability of ecological assessments.

Moreover, our analysis of the most advanced click-based interactive segmentation models provides valuable insights into their performance and applicability in remote sensing. One notable observation is that the SimpleClick-CFR and ICL-CFR models have consistently performed best across various evaluation metrics (Tables 4-8). This indicates that the CFR strategy integrated in this study notably improved the models' performance. Bytedance successfully applied the CFR approach to enhance the COCO dataset in 2024 (Deng et al., 2024). The CFR strategy is advantageous for remote sensing images due to its iterative refinement capability, which incrementally enhances segmentation accuracy essential for capturing fine details in complex datasets. By integrating user interaction, such as click-based inputs, CFR addresses ambiguities and improves initial segmentation, which is crucial in differentiating subtle features in high-resolution images. Its adaptability in setting refinement steps optimizes processing for varying image complexities, improving efficiency without repetitive user input. This approach is particularly beneficial for applications like land use classification and environmental monitoring, where precision and robustness against variability in image quality are paramount.

We then focus on the impact of different band combinations on the model performance. Including the IR band improved vegetation monitoring and analysis (Tables 5 and 7). While the inclusion of the IR band was expected to enhance vegetation identification, our experiments using RGB and RGIR band combinations indicate that the models' performance on both PRGB and PRGIR datasets remains comparable (Tables 6 and 7). This suggests that the models can effectively segment vegetation regardless of the specific band combination. Beyond vegetation, we also analyzed other land cover classes, such as buildings, impervious surfaces, and cars. Among these classes, buildings demonstrated the most favorable results, as their regular and well-defined shapes facilitated accurate segmentation.

Lastly, we extended our analysis to explore the association between land cover object sizes and model segmentation performance. By systematically evaluating how different sizes of land objects, such as buildings and trees, affect the number of input points needed to achieve a desired Intersection over Union (IoU), we aimed to optimize the efficiency and accuracy of the segmentation process. Our results indicate that smaller land objects require more clicks for accurate segmentation (Table 9). This suggests that the fine details and intricate structure of smaller objects require additional user input for precise delineation.

Overall, the SimpleClick-CFR and ICL-CFR models exhibit promising performance in remote sensing applications, effectively segmenting various land cover classes, including vegetation and buildings. Additionally, these findings highlight the influence of object size on the segmentation process and provide insights into the scalability of these methods to regions with varying feature densities and complexities.

5.2 SegMap Tool

Several geo-based SAM tools have been previously proposed, such as the SAM API for ArcGIS Pro ("Introduction to the model—ArcGIS pretrained models | Documentation," n.d.), SAMGEO, a Python package for a convenient application of SAM on spatial data (Wu and Osco, 2023) and SAMJS, a SAM API based geographical classification tool developed by AntV. However, most lack interactivity and efficiency, limiting their application to interactive segmentation tasks. To address this gap, we developed a highly interactive plug-in, SegMap, for the free-access and open-source platform QGIS, to perform deep learning-based interactive segmentation on remote sensing data. This tool offers several key features, as outlined below:

1. SegMap is a QGIS plugin leveraging deep learning to facilitate rapid map digitization. It enables users to efficiently extract objects from aerial imagery with minimal effort, streamlining the process and replacing traditional, labor-intensive polygon editing methods.
2. SegMap significantly enhances productivity by allowing users to complete tasks in seconds with just 1-5 clicks, compared to the traditional method that involves painstakingly adjusting hundreds of vertices. This drastic improvement in efficiency means that complex map digitization tasks, which would generally take considerable time and effort, can now be completed swiftly and with minimal manual intervention.
3. The plugin harnesses the power of advanced deep learning models to achieve precise object extraction from aerial imagery. By utilizing sophisticated algorithms, SegMap is capable of accurately identifying and delineating objects, ensuring high-quality results. This AI-driven approach not only improves accuracy but also reduces the potential for human error, making it a valuable tool for users requiring quick and reliable digitization of maps.

Looking ahead, we plan to further optimize SegMap by enhancing its capabilities for automated segmentation and incremental learning, ultimately improving land use classification from remote sensing imagery, and further developing an ArcGIS Pro toolbox.

**6. Conclusion**

This paper evaluates the robustness and efficiency of the latest interactive segmentation models for remote sensing applications. By conducting an extensive analysis across various remote sensing datasets, we examined the influence of different band combinations, object sizes, and object categories. Our results show that segmentation can be achieved with as few as two points for larger and simpler objects, such as buildings over 160 $m^2$. Additionally, by evaluating different models and comparing them with the widely used SAM model, we identified SimpleClick as the most efficient and robust model for click-based interactive segmentation.

Based on these findings, we developed SegMap, an interactive segmentation plug-in under QGIS tailored for remote sensing imagery. Utilizing the ICL-CFR and SimpleClick-CFR models fine-tuned with the remote sensing datasets, the segmented output results can be further refined and analyzed conveniently using the QGIS platform, enhancing the tool's applicability in geospatial analysis.

Looking ahead, several improvements can be made based on our findings. We plan to use SegMap to create additional training samples to improve the model's accuracy in classifying remote-sensing images.

Furthermore, there is great potential to enhance the ICL-CFR model, adapting it to the complexity of remote sensing data. The CFR process can be repeated to enable the model to learn more from previously obtained results. Additionally, the refinement process could be modified to use the information from previous masks fully. In summary, the uniqueness of remote sensing images allows for detailed analysis in both interactive segmentation and deep learning aspects.

**Code and dataset availability**

The code for SegMap tool is available at https://github.com/TitorX/SegMap-QGIS.

The code for CFR-ICL is available at https://github.com/TitorX/CFR-ICL-Interactive-Segmentation

Contact: Shoukun Sun (sun5322@vandals.uidaho.edu) or Xiang Que (xiangq@uidhao.edu)

Software Platform: QGIS 3.40-Bratislava LTR or more recent version.

Hardware requirements: CPU, AMD Ryzen 9 3900X or better. RAM, at least 16 GB. GPU, NVIDIA GeForce RTX 3090 or higher.

Program language: Python and TypeScript

Program size: At least 2GB of disk space.

**Funding**

This work is supported by the National Natural Science Foundation of China under Grant No.42202333.

**CRediT authorship contribution statement**

Z.Wang: Conceptualization, Data curation, Formal analysis, Investigation, Methodology, Visualization, Writing – original draft, Writing – review & editing. S.Sun: Formal analysis, Visualization, Methodology – review & editing. X.Que: Supervision, Validation, Writing-review & editing. X.Ma: Writing – review & editing. C. Galaz García: Review & editing

**Declaration of Competing Interest**

The authors declare that they have no known competing financial interests or personal relationships that could have appeared to influence the work reported in this paper.